\pdfoutput=1

\documentclass[11pt]{article}

\usepackage{emnlp2021}

\usepackage{times}
\usepackage{url}
\usepackage{caption}
\usepackage{graphicx}
\usepackage{amsmath}
\usepackage[normalem]{ulem}
\usepackage{booktabs}
\usepackage{enumitem}
\usepackage{microtype}
\usepackage{booktabs}
\usepackage{eqnarray}
\usepackage{amsfonts}
\usepackage{latexsym}
\usepackage{subfigure}
\usepackage{algorithm}
\usepackage{algorithmicx}
\usepackage{multirow}
\usepackage{xifthen}
\usepackage{diagbox}
\usepackage{pfnote}
\usepackage{bm}
\usepackage{xcolor}
\usepackage{dsfont}
\usepackage{xspace}

\usepackage[T1]{fontenc}

\usepackage[utf8]{inputenc}

\usepackage{microtype}

\newcommand{\ccqg}[0]{CCQG\xspace}
\title{Simple or Complex? Complexity-Controllable Question Generation with Soft Templates and Deep Mixture of Experts Model}

\author{Sheng Bi$^{1}$\and
Xiya Cheng$^{1}$\and
Yuan-Fang Li$^{2}$\and
Lizhen Qu$^{2}$\and
Shirong Shen$^{1}$\\
\textbf{Guilin Qi}$^{1*}$\and
\textbf{Lu Pan}$^{3}$\and
\textbf{Yinlin Jiang}$^{1}$
 \\
$^{1}$School of Computer Science and Engineering, Southeast University, China \\
$^{2}$Faculty of Information Technology, Monash University, Melbourne, Australia \\
$^{3}$Baidu Inc., China \\
\tt\{bisheng,chengxiya\}@seu.edu.cn,\{yuanfang.li,lizhen.qu\}@monash.edu \\ \tt  \{ssr,gqi\}@seu.edu.cn,panlu01@baidu.com,yljiang@seu.edu.cn}

\begin{document}
\maketitle
\begin{abstract}
The ability to generate natural-language questions with controlled complexity levels is highly desirable as it further expands the applicability of question generation. 
In this paper, we propose an end-to-end neural complexity-controllable question generation model, which incorporates a mixture of experts (MoE) as the selector of soft templates to improve the accuracy of complexity control and the quality of generated questions. The soft templates capture question similarity while avoiding the expensive construction of actual templates.
Our method introduces a novel, cross-domain complexity estimator to assess the complexity of a question, taking into account the passage, the question, the answer and their interactions.
The experimental results on two benchmark QA datasets demonstrate that our QG model is superior to state-of-the-art methods in both automatic and manual evaluation. Moreover, our complexity estimator is significantly more accurate than the baselines in both in-domain and out-domain settings. 
\end{abstract}
\abovedisplayskip=0pt
\abovedisplayshortskip=0pt
\belowdisplayskip=0pt
\belowdisplayshortskip=0pt
\abovecaptionskip=0pt
\belowcaptionskip=0pt

\section{Introduction}
The task of Question Generation (QG) aims at generating natural-language questions from different data sources, including passages of text, knowledge bases, images and videos. For a variety of applications, it is highly desirable to be able to \emph{control} the complexity of generated questions. 
For instance, in the field of education, a well-balanced test needs questions of varying complexity levels in suitable proportions for students of different levels~\citep{DBLP:conf/owled/AlsubaitPS14}. That is to say, the teacher can tailor the questions to the competence of the learner. In addition, it has recently been shown~\citep{DBLP:conf/acl/SultanCAC20} that Question Answering (QA) models can benefit from training datasets enriched by applying QG models. However, despite the growing interests of answering complex questions~\citep{DBLP:conf/naacl/CaoAT19} 
as well as questions with varying complexity levels~\citep{seyler2017knowledge}, most existing work focus on generating simple questions~\citep{DBLP:conf/nlpcc/ZhouYWTBZ17}. Although Pan et al.~\citeyearpar{DBLP:conf/acl/PanXFCK20} explored the generation of complex questions, they do not consider controlling the complexity of generated questions. 
Complexity-controllable question generation (CCQG) faces a number of challenges. 

\textbf{High diversity.} Compared to simple questions, complex questions contain significantly more information and exhibit more complex syntactic structures. The complexity of questions is caused by compositional complexity because complex questions can be decomposed to a sequence of simple questions~\citep{DBLP:conf/emnlp/PerezLYCK20}. Generation of both simple and complex questions imposes even higher challenges because simple and complex questions demonstrate different semantic and syntactic patterns. To this end, the resulted distributions are expected to be multimodal, i.e., with different modes for different patterns of questions.

Existing works~\citep{gao2019difficulty,kumar2019difficulty} fail to capture the diverse nature of \ccqg. They model complexity as discrete labels, such as \emph{easy} and \emph{hard}, and introduce a learnable embedding as the representation of the complexity labels in the initial hidden state at the decoding stage. However, the information contained in such an embedding plays a limited role in modelling multiple modes of the underlying distribution. Similarly, it is observed that latent variables are ignored such that the posterior is always equal to the prior in variational autoencoders~\citep{DBLP:conf/conll/BowmanVVDJB16}.

\textbf{Limited training data.} The training of \ccqg models requires questions annotated with complexity levels. However, although there are a large number of QA datasets in various domains, few of them is annotated with complexity levels. Therefore, in-domain training of high quality \ccqg models becomes infeasible in most domains.

In this paper, we propose a novel question generation model, \ccqg, capable of controlling question complexity. We incorporate soft templates and deep \emph{mixture of experts} (MoE) \cite{DBLP:conf/icml/ShenOAR19} to address the high diversity problem. Inspired by a recent work~\citep{cao-etal-2018-retrieve}, we posit that similar questions have similar templates, and that different modes of the underlying distributions should capture different question templates. Instead of manually constructing templates, which is labor-intensive and time-consuming, we employ \textbf{soft templates}, each of which is a sequence of latent embeddings. 
Inspired by Cho et al~\citeyearpar{DBLP:conf/emnlp/ChoSH19}, we apply MoE to select templates, whereby we introduce a discrete latent variable to indicate the choice of an expert. Taking as input a complexity level, a passage and an answer, our model selects an expert, which chooses a 
template of that complexity level to guide the question generation process. 

To address the challenge of limited training data, we design a simple and effective cross-domain complexity estimator based on five domain-independent features to classify questions w.r.t.\ their complexity levels. The predicted labels are incorporated into the training of \ccqg.  
The main contributions of this work are three-folds:

\begin{itemize}[leftmargin=*,nosep]
	\item An end-to-end neural complexity-controllable QG model, which incorporates mixture of experts (MoE) and soft templates to model highly diverse questions of different complexity levels.
	\item  A simple and effective cross-domain complexity estimator to assess the complexity of a question. 
	\item We evaluate our CCQG model and complexity estimator on two benchmark QA datasets, SQuAD~\citep{DBLP:conf/acl/RajpurkarJL18} and HotpotQA~\citep{DBLP:conf/emnlp/Yang0ZBCSM18}. The experimental results demonstrate that our QG model is superior to baselines in both automatic and human evaluation. The complexity estimator significantly outperforms the strong baselines with pre-trained language models in both in-domain and out-domain settings. The source code will be released to encourage reproducibility.
\end{itemize}

\section{Related Work}
Our work is mainly relevant to question complexity estimation and question generation.
\subsection{Question Complexity Estimation}
Several methods have been proposed to determine the complexity of questions.
~\citep{DBLP:conf/owled/AlsubaitPS14} presented a similarity-based theory to control the complexity of multiple-choice questions and showed its consistency and efficiency with educational theories. \citep{seyler2017knowledge,kumar2019difficulty} estimated the complexity of questions with similar manner. In general, they made statistical analysis on some features of the entities in the question, such as popularity, selectivity, and coherence, so as to evaluate the complexity. 
These estimation methods only focus on the questions themselves while ignoring the effect of the associated input context. Intuitively, a question has distinct complexities with different contexts.

Gao et al.~\citeyearpar{gao2019difficulty} evaluated the difficulty levels of questions in datasets based on whether reading comprehension systems can answer or not. This method relies heavily on the quality of QA systems and is not accurate enough.
For a learner (human or machine), there are typically three iterative steps involved in answering a question, reading the passage, understanding the question, and finding the answer, which means that the complexity of a question should consider these three parts. 
\subsection{Question Generation}
The existing work of question generation (QG) can be roughly divided into two directions, rule-based and neural-based. The former~\citep{heilman2011automatic} usually relies on manually designing lexical rules to generate questions, which is labor-intensive and has poor scalability.
With the success of deep learning, many sequence-to-sequence (Seq2seq) models have been proposed for QG tasks. \citep{DBLP:conf/nlpcc/ZhouYWTBZ17} used enriched semantic and lexical features in QG with attention and copy mechanism~\citep{DBLP:conf/acl/SeeLM17}. 
~\citep{DBLP:conf/coling/BiCLWQ20} designed a new reward with grammatical similarity to improve the syntactic correctness of generated question through reinforcement learning.

Due to the demand for different complexity-level questions in real scenarios, researchers began to explore generating complexity-controllable questions. ~\citep{kumar2019difficulty} used named entity popularity to estimate difficulty and generated difficulty-controllable questions. Besides, ~\citep{gao2019difficulty} evaluated the difficulty levels of questions based on QA systems and generated questions under the control of specified difficulty labels. These two models are similar in that they encode the complexity labels and use the encoded vectors as the complexity-controllable constraint. Due to the lack of parallel corpus in real scenes, which means there is only one question with ``simple'' or ``complex'' level for a pair of passage and answer. Only relying on one vector as a condition for controlling complexity, it is difficult to make the generated question conform to the given complexity constraint.

Therefore, in this paper, we propose an adaptive, generalizable complexity evaluator that considers both the question and the context while evaluating the question complexity independent of any QA system.
In addition, we propose a novel model of CCQG. Compared to traditional methods that encoding complexity with only single vector as complexity constraint, we introduce mixture of experts~\citep{DBLP:conf/emnlp/ChoSH19} to ensure the diversity of questions with different complexity levels. We also introduce the soft template to improve the fluency of the generated questions.

\section{Methodology}

Given a passage, an expected answer, and a complexity level, the task of \ccqg is to generate questions with the specified complexity. According to~\cite{kunichika2002computational}, the complexity of a question depends on two factors: i) individual capability of answering a question, and ii) the common process required to answer a question (e.g.\ understanding content of a question and background knowledge, steps of reasoning to infer an answer). The former varies between individual learners so that it is infeasible to find a generally applicable criterion. 
Despite that, we can determine the shared factors involved in the answering process and use them to quantify complexity of a question. The resulted score is then used to categorize complexity of a question. More details can be found in Sec. \ref{Complexity Computational Method}. 

Formally, given a passage denoted as a word sequence $X =(x_1,\cdots,x_{n_X})$ with $x_i$ in a vocabulary $\mathcal{V}$, a complexity level $d \in \{\text{simple}, \text{complex}\}$, an answer $A= (x_1,\cdots,x_{m})$, our goal is to generate the most probable question $\hat{
Y}=(y_1,\cdots,y_{n_Y})$ with $y_i \in \mathcal{V}$, which has $A$ as its answer and the complexity level $d$. The estimation of complexity level $d$ will be described in Sec.~\ref{Complexity Computational Method}. 
\begin{align}
    \hat{Y}&=\mathop{\arg\max}\limits_{Y}p(Y|X,A,d).
    \label{QG}
\end{align}
\begin{figure}
\centering
    \includegraphics[width=0.45\textwidth]{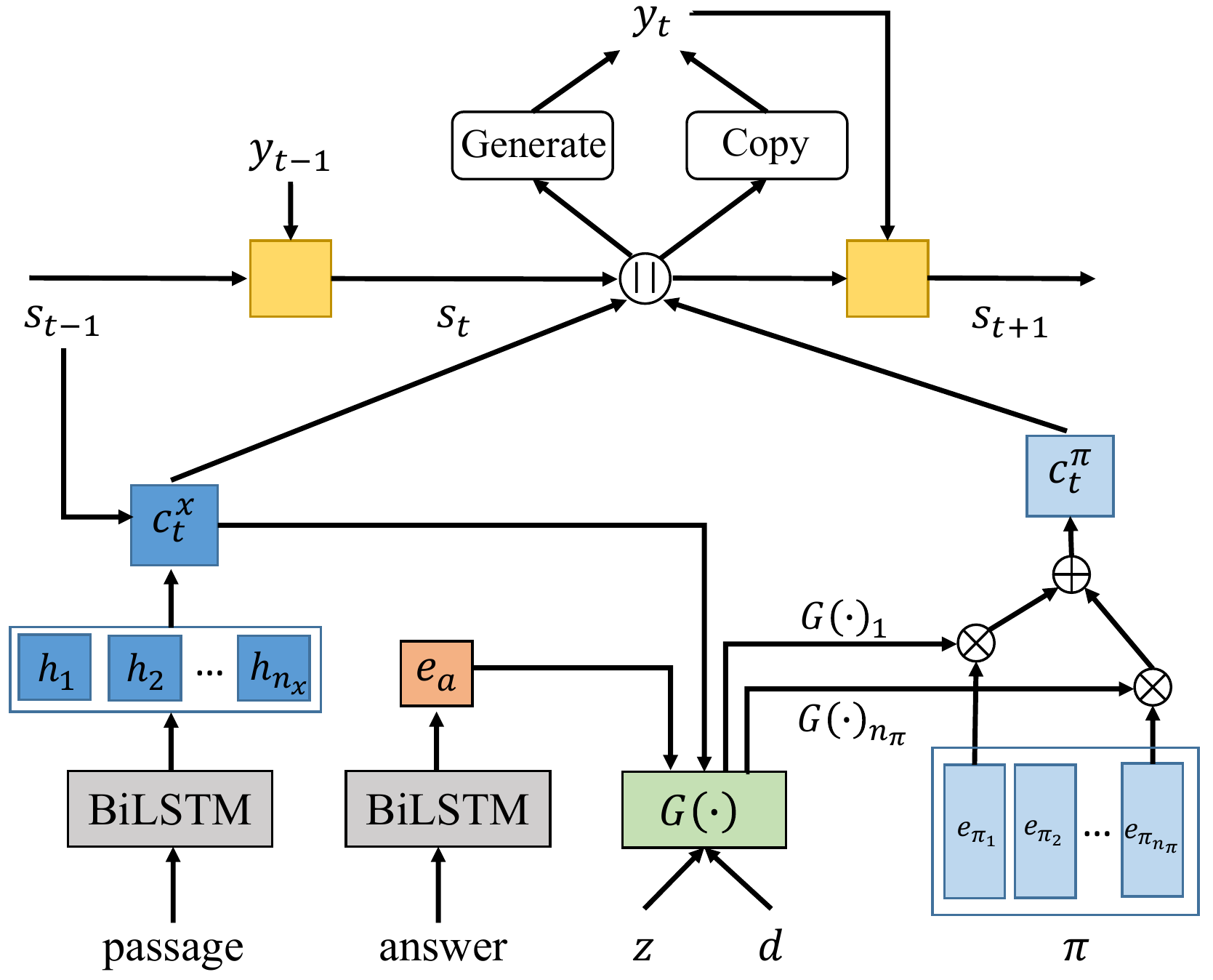}
    \caption{The overall framework of our \textbf{CCQG} model. CCQG consists of four main modules: (1) BiLSTM-based encoders of passage and answer (gray); (2) MoE-based template element selector for inputting experts and complexity and outputting probability distributions for different templates (green); (3) template element representation blocks initialized by the centroids of the question clusters at the corresponding complexity (light blue); (4) conditioned question generator (yellow).}
    \label{model}
\end{figure}
Given the same passage, there are different ways of asking questions, which can be summarized into different \emph{question templates} that model their semantic similarities~\citep{cao-etal-2018-retrieve} and complexity similarities. Templates provide a reference point as guidance for more nuanced question generation.
~\citet{cao-etal-2018-retrieve} also suggested that questions generated from templates tend to be fluent and natural. Therefore, we argue that question generation would be more effective if the model chooses the appropriate templates at each decoding step. 

Despite the usefulness of templates for question generation, template construction is labor-intensive and requires substantial domain knowledge. Therefore, template-based QG approaches typically suffer from low coverage.  
To alleviate this problem, we employ \textbf{soft templates} and avoid explicitly designing string-based templates. A soft template is modeled as a sequence of elements, each of which provides a reference point at a decoding step $t$. This modeling allows sharing of elements across templates at the same complexity level. The selection of soft templates is conducted through a mixture of experts. Each expert distinguishes from each other in terms of its preference of templates. For a given input, an expert from them is chosen to determine a probable template. 
Both soft templates and experts are latent. As a template is a sequence of template elements, we introduce a latent variable $\pi^d_t \in \{1,\cdots,n_{\pi}\}$ for the selection of template elements at the complexity level $d$ at time $t$. The value of $\pi^d_t$ indicates the choice of a particular template element. In the same manner, we introduce another latent variable $z \in \{1,\cdots,n_z\}$ to represent the choice of an expert for a given input. Each expert has its own dense vector representation $\mathbf{e}_z$. 
At each decoding step $t$, we obtain the probability of estimating $y_t$ by marginalizing over all template elements. We also marginalize over all latent experts for the same input.

\begin{align} \nonumber
    p(Y|X,A,d) \\= \sum_{z=1}^{n_z}\prod_{t=1}^{n_Y} \sum_{\pi^d_t = 1}^{n_{\pi}}& [p(y_t|y_1,\dots,y_{t-1},X,A,\pi^d_t) \nonumber \\
    & p(\pi^d_t | X, A, z)]p(z|X,A,d),
    \label{template marginallize}
\end{align}
where $p(z|X,A,d)=1/n_z$ is the uniform prior probability of the experts, because it has been observed~\citep{DBLP:conf/icml/ShenOAR19} that the uniform prior encourages the model to make use of all the components for each input context. The control of complexity is achieved by choosing the set of possible template elements through $d$ and an expert $z$ is chosen to select a probable soft template based on a given input. 

\subsection{Model Details}
As shown in Figure \ref{model}, our model consists of a passage encoder, an answer encoder and a question decoder. Each encoder employs a Bidirectional LSTM (BiLSTM)~\citep{hochreiter1997long} with different parameterizations, respectively.  
The question decoder is modeled by using a single layer LSTM with soft attention~\citep{DBLP:journals/corr/BahdanauCB14} and a softmax layer. 

The LSTM decoder utilizes soft templates and mixture of experts to generate complexity-controllable questions. As input, it takes the previous generated word $y_{t-1}$, the current context vector $\mathbf{c}_t^x$, the aggregated representation of the soft template elements $\mathbf{c}^{{\pi_t^d}}$, the embedding of an expert $\mathbf{e}_z \in \mathbb{R}^{d_z}$, and the previous hidden state $s_{t-1}$.
\begin{align}
\label{eq:decoder}
\mathbf{s}_t=LSTM(fc([\mathbf{y}_{t-1},\mathbf{c}_t^x,\mathbf{c}_t^{\pi},\mathbf{e}_z]),\mathbf{s}_{t-1}),
\end{align}
where $fc$ denotes a full connected layer. The current context vector $\mathbf{c}_t^x$ is created by attending over the hidden representations of the passage encoder, following ~\citep{DBLP:journals/corr/BahdanauCB14}. We initialize the first hidden state $s_0$ as $fc([h_{n_x},\mathbf{e}_a,d,\mathbf{e}_z])$, where $\mathbf{e}_a$ denotes the embedding of the input answer $a$.

The soft template representation $\mathbf{c}_t^{\pi}$ at time $t$ is aggregated over all template elements at the complexity level $d$, which is calculated as 
\begin{align}
	\mathbf{c}_t^{\pi}=\sum_{i=1}^{n_{\pi^d}}p(\pi^d_i|\mathbf{c}_t^x, X,A,d,z)\mathbf{e}_{\pi^d_i},
\end{align}
where $\mathbf{e}_{\pi^d_i}$ denotes the trainable embedding of an element $\pi^d_i$. The module $p(\pi_i|\mathbf{c}_t^x, X,A,d,z)$ estimates the relevance of a template element at time $t$. We consider soft attention over hard attention because it allows more than one elements to be relevant to the current context and the input. We take $p(\pi|X,A,d,z)$ as a learned prior distribution and model it with a gating network $G(\cdot)$. Moreover, to encourage sparse selection of elements, we model $G(\cdot)$ with choosing only the top-$k$ most relevant ones by applying the noisy $\operatorname{TopK}$ gating network~\citep{DBLP:conf/iclr/ShazeerMMDLHD17}. This network also helps load balancing by introducing a noise term. More details can be found in~\citep{DBLP:conf/iclr/ShazeerMMDLHD17}. 
As a result, we obtain $\mathbf{c}_t^{\pi}$ by:
\begin{align}
	\mathbf{c}_t^{\pi}=\sum_{i=1}^{K}\text{softmax}(\operatorname{TopK}([\mathbf{c}_t^x,\mathbf{e}_a,d,\mathbf{e}_z]))\mathbf{e}_{\pi^d_i}.\nonumber
\end{align}

The parameters of each expert embedding $\mathbf{e}_z$ are initialized randomly and fine-tuned during training. During decoding, we iterate through all experts to generate $n_z$ question candidates. Among them, the question with the highest $p(Y|X,A,d,z)$ is chosen as the final prediction.

Each state $\mathbf{s}_t$ in Eq.\eqref{eq:decoder} is fed to the pointer-generator network~\citep{DBLP:conf/acl/SeeLM17} for generation of each word. This module is chosen to overcome out-of-vocabulary (OOV) words by coping them from input passages on demand. 

\subsection{Training}
During training, we initialize the template elements by using questions at the respective complexity level and train the whole model with hard EM.
\paragraph{Template Element Initialization} We initialize the embeddings of template elements by using the centroids of the question clusters at the respective complexity level. Compared to random initialization, it encourages embeddings to capture the intrinsic properties of distinct question templates. More specifically, we encode each question in the train set by using BERT~\citep{devlin-etal-2019-bert}. Then we cluster the outcomes at each complexity level by using the improved k-means algorithm~\citep{DBLP:conf/iitsi/ShiLG10}. The resulted cluster centroids are taken as the initial embeddings. 
\paragraph{Training with Hard-EM} We train the model with hard-EM~\citep{DBLP:conf/nips/LeePCRCB16} by taking the following two steps iteratively until convergence, because hard-EM can learn more diverse experts than soft-EM in NLG tasks~\citep{DBLP:conf/icml/ShenOAR19}. 

\noindent\textbf{E-step (hard).} We calculate the loss for each expert and choose the expert with the minimal loss as the best one $z^*$. 
\begin{align}\nonumber
	z^*&=\mathop{\arg\min}\limits_{z}-\log p(Y|X,A,d,z).
\end{align}
\noindent\textbf{M-step.} We optimize the model parameters $\theta$ with the best expert $z^*$.
\begin{align}\nonumber
\mathop{\min}\limits_{\theta}-\log p(Y|X,A,d,z^*;\theta).
\end{align}

\section{Cross-Domain Complexity Estimator}
\label{Complexity Computational Method}
It is desirable to build a cross-domain estimator to predict the complexity levels of questions because few domains have questions annotated with complexity levels for training \ccqg models. As measuring complexity should be independent of domain-specific content, we use a simple classification rule without any training, which relies on the following five domain-independent features $d_{f_i}$. 

\paragraph{Number of clauses in a question ($\bm{d_{f_1}}$):} The number of events/facts is a strong indicator of question complexity. We observe that the number of clauses are often proportional to the number of events/facts mentioned in a question. We use NLTK\footnote{\url{http://www.nltk.org/}} to seek the question's syntactic tree to count the number of clauses.

\paragraph{Number of certain dependency relations in a question ($\bm{d_{f_2}}$):} Certain dependency relations across words influences the understanding of the content of a question~\citep{kunichika2002computational}. The more of them, the more difficult it is to understand.
Thus, we count the number of \textit{advmod, amod, nounmod, npmod}, and possessive modifiers after running the Spacy dependency parser~\footnote{\url{https://spacy.io/}} on questions.

\paragraph{Topic coherence of sentences in a passage ($\bm{d_{f_3}}$):} Kunichika et al~\citeyearpar{kunichika2002computational} observed that a passage is easy to understand if the topic coherence of its sentences is high. In light of this, we measure the topic coherence between sentences by calculating the Jensen–Shannon Divergence $\mathcal{JS}$~\citep{MENENDEZ1997307} between their topic distributions.
$
    d_{\mathcal{JS}}=\frac{1}{n(n-1)}\sum_{i\neq j}\mathcal{JS}(\mathbf{t}_i,\mathbf{t}_j),
$
where $n$ is the number of sentences in a passage, and $\mathbf{t}_i$ and $\mathbf{t}_j$ denote the topic distributions of the $i$-th and $j$-th sentences in a passage respectively. 
As we expect the feature value is high if a question is complex, we let this feature $d_{f_3} = 1/ d_{\mathcal{JS}}$.

\paragraph{Frequency of question entities in a passage ($\bm{d_{f_4}}$):} We observe that a question asking about an entity frequently appearing in a passage is often easier to answer than the one about an infrequent entity. 
Thus, we recognize entities in questions and passages, compute the average frequency of entities mentioned both in a question and a passage by $\text{avg}(Q) = \frac{1}{|E^Q|} \sum_{E^Q}\frac{n_{e_i}}{\sum_{E^P}n_{e_j}}$, where $E^Q$ denotes the entity set in the question, $E^P$ denotes the entity set in the passage and $n_{e_i}$ is the number of mentions of $e_i$ in the passage. Then the feature is the inverse of the averaged frequency $d(f_4) = 1/\text{avg}(Q)$. 

\paragraph{Distance between entities in a question and an answer span in a passage ($\bm{d_{f_5}}$):} The answer to a question is often easy to find, if an entity mentioned in the question is located close to the answer in the same passage. Therefore, $d_{f_5}$ is such a distance by taking the average number of tokens between the entities in a question and an answer span in a passage.

\paragraph{Classification rule} The scoring function based on the above features is the average of all feature values after normalization $cpx(Q)=\frac{1}{5}\sum_{i=1}^{5}\text{Norm}(d_{f_i}(Q))$, where $ \text{Norm}(d_{f_i}(Q)) =\frac{d_{f_i}(Q)-\min(d_{f_i}(Q))}{\max(d_{f_i}(Q))-\min(d_{f_i}(Q))}$. We consider a question $Q$ as \textit{complex}, if $cpx(Q)$ is above a threshold $\lambda$, otherwise the question is classified as \textit{simple}. The threshold can be easily tuned on a small sample of data annotated with complexity levels.

\section{Experiments}
In this section, we evaluate the effectiveness of the CCQG model and the complexity estimator.

\subsection{Datasets and Complexity Annotation}
We conduct experiments on two benchmark datasets SQuAD~\citep{rajpurkar-etal-2016-squad} and HotpotQA~\citep{DBLP:conf/emnlp/Yang0ZBCSM18}. We remove the questions that are unanswerable or whose answers are not contiguous fragments in the passage. For each dataset, we randomly select 80\% of samples for training, 10\% for validation, and 10\% for testing. 

We use only predicted complexity levels for training \ccqg models on both datasets. In particular, we apply the cross-domain estimator to label each question with complexity levels. We calibrate the threshold $\lambda$ on the questions labeled by \textit{easy} and \textit{hard} in the train set of HotpotQA, because only the questions in HotpotQA contain manually annotated complexity levels. The resulted $\lambda = 0.682$ is used in both HotpotQA and SQuAD. 
Table~\ref{dataset} summarizes the data statistics. 

\begin{table}[htp]
  \centering
  \caption{The statistics of HotpotQA and SQuAD.}
  \label{dataset}
\resizebox{0.4\textwidth}{!}{
  \renewcommand\tabcolsep{2pt} 
  \begin{tabular}{lcccccc}
    \toprule
     & \multicolumn{3}{c}{HotpotQA} & \multicolumn{3}{c}{SQuAD}  \\
     \cmidrule(lr){2-4} \cmidrule(lr){5-7} 
    
    &  train     &   dev  &   test
    &  train     &   dev  &   test  \\
    \midrule
   \textit{simple} & 45,585 & 5,698 & 5,426 & 41,604 & 5,201  & 5,235 \\
    \midrule
    \textit{complex} & 26,772  & 3,346 & 3,617 &27,852 & 3,386 & 3,446 \\
    \bottomrule
  \end{tabular}
  }
\end{table}

\begin{table}[htb]
\centering
\caption{Results of automatic evaluations on SQuAD and HotpotQA for varying complexity levels, the best performance is in bold.}
  \label{auto results on datasets}
\resizebox{0.45\textwidth}{!}{
  \begin{tabular}{lcccccc}
    \toprule
    Datasets& \multicolumn{3}{c}{SQuAD-simple} & \multicolumn{3}{c}{SQuAD-complex} \\
     \cmidrule(lr){2-4} \cmidrule(lr){5-7} 
    Metrics
    &  B-4      &   R-L  &   F1 &  B-4      &   R-L  &   F1  \\
    \midrule
    
   NQG++      & 12.19  & 45.39 & 69.26 & 11.16  & 43.70 & 65.39  \\
   DLPH       & 12.65  & 46.01 & 70.15 & 10.91  & 45.43 & 67.01  \\ 
   DeepQG     & 15.50  & 54.05 & 70.22 & 14.25  & 52.13 & 71.53  \\
   MoE        & 12.55  & 46.52 & 71.85 & 12.38  & 45.58 & 69.11  \\ 
   \midrule
  CCQG        & \textbf{17.14} & \textbf{54.28} & \textbf{78.60} & \textbf{16.01} & \textbf{53.19} & \textbf{74.81} \\
    \midrule
    w/o z     & 16.02  & 52.13 & 75.25  & 14.95  & 51.08 & 72.40 \\
    w/o $\pi$ & 13.05  & 46.21 & 72.81  & 12.57  & 44.19 & 69.37 \\
    \bottomrule
  \end{tabular}
  }

\resizebox{0.45\textwidth}{!}{
  \begin{tabular}{lcccccc}
    \toprule
    Datasets& \multicolumn{3}{c}{HotpotQA-simple} & \multicolumn{3}{c}{HotpotQA-complex} \\
     \cmidrule(lr){2-4} \cmidrule(lr){5-7} 
    Metrics
    &  B-4      &   R-L  &   F1 &  B-4      &   R-L  &   F1  \\
    \midrule
    
   NQG++      & 12.35  & 44.51 & 63.37 & 10.76 & 41.26 & 64.05 \\
   DLPH       & 12.01  & 43.28 & 68.98 & 11.50 & 43.58 & 66.71 \\ 
   DeepQG     & 14.25  & 50.18 & 67.25 & 13.66 & 49.17 & 68.86 \\
   MoE        & 12.95  & 44.31 & 72.83 & 11.68 & 43.20 & 68.19 \\ 
   \midrule
  CCQG        & \textbf{17.85}  & \textbf{55.36}& \textbf{80.57}& \textbf{15.41}  & \textbf{53.73}& \textbf{76.19} \\
    \midrule
    w/o z     & 16.73 & 53.07 &77.12 & 14.26 & 51.85 &74.70 \\
    w/o $\pi$ & 13.87 & 46.98 &73.91 & 13.01 & 46.50  &70.05 \\
    \bottomrule
  \end{tabular}
  }
\end{table}

\begin{table}[htp]
  \centering
  \caption{Results of human evaluations on SQuAD and HotpotQA with different complexity levels, the best performance is in bold.}
  \label{human results on datasets}
\resizebox{0.48\textwidth}{!}{
  \renewcommand\tabcolsep{2.4pt} 
  \begin{tabular}{lcccccccccc}
    \toprule
    Datasets& \multicolumn{2}{c}{SQuAD-} & \multicolumn{2}{c}{SQuAD-} &All& \multicolumn{2}{c}{HotpotQA-}&\multicolumn{2}{c}{HotpotQA-} & All\\
            & \multicolumn{2}{c}{simple} & \multicolumn{2}{c}{complex} & & \multicolumn{2}{c}{simple}&\multicolumn{2}{c}{complex} & \\
     \cmidrule(lr){2-3} \cmidrule(lr){4-5} \cmidrule(lr){6-6} \cmidrule(lr){7-8} \cmidrule(lr){9-10}\cmidrule(lr){11-11} 
    Metrics
    & \;\; Nat.  &   Cpx.  &  \;\;\; Nat.   &Cpx. & Div.
    & \;\;\; Nat.  &   Cpx.  &  \;\;\;\; Nat. &Cpx. & Div.  \\
    \midrule
    
   NQG++ &  \;\; 2.6 & 2.6 &\;\;\;2.5 & 2.4 & 2.1 & \;\;\;2.6  & 2.9 & \;\;\;\;2.3  & 2.4 & 1.9 \\
   DLPH  & \;\;  2.7 & 2.5 & \;\;\;2.7 & 2.7 & 2.7 & \;\;\;2.6  & 2.7 & \;\;\;\;2.5 & 2.7 & 2.4 \\ 
   DeepQG &  \;\; 3.1 & 2.2 & \;\;\;3.0 & 2.9 & 2.3 & \;\;\;2.9  & 2.5 & \;\;\;\;3.0 & 2.9 & 2.1 \\
   MoE &  \;\; 2.9 & 1.9 & \;\;\;2.9 & 2.9 &2.9 & \;\;\;3.1  & 2.3 & \;\;\;\;2.8 & 2.7 & 2.8 \\ 
   \midrule
  CCQG&  \;\; \textbf{3.6} &\textbf{1.5} & \;\;\;\textbf{3.5} &\textbf{3.3} & \textbf{3.6} & \;\;\;\textbf{3.7}  &\textbf{1.3}& \;\;\;\;\textbf{3.6}  &\textbf{3.4}&\textbf{3.6} \\
    \midrule
    w/o z & \;\; 3.5 & 1.7  &\;\;\;3.3 & 3.1 &3.0 & \;\;\;3.6   & 1.6 & \;\;\;\;3.3    & 3.2 &3.4 \\
    w/o $\pi$ & \;\; 3.0  & 1.9  &\;\;\;2.8 & 2.9 & 2.9 & \;\;\;3.0     & 2.2  & \;\;\;\;2.8     &2.6  & 2.9\\
    
    \bottomrule
  \end{tabular} 
  }
\end{table}

\subsection{Settings for CCQG}
\paragraph{Baselines.}
We compare our models with the following baselines on the two datasets.

\noindent{\textbf{NQG++:}} an encoder-decoder model with attention and copy mechanisms for QG tasks. It introduces lexical features and the answer position to enhance semantic representation~\citep{DBLP:conf/nlpcc/ZhouYWTBZ17}. 

\noindent{\textbf{DLPH:}} an end-to-end difficulty-controllable QG model,  which estimates the complexity level of a question based on whether the QA systems can answer it correctly or not~\citep{gao2019difficulty}.

\noindent{\textbf{DeepQG:}} an attention-based gated graph neural network that fuses the semantic  representations of document-level and graph-level to select content and generate complex questions~\citep{DBLP:conf/acl/PanXFCK20}.

\noindent{\textbf{MoE:}} a method for diverse generation that uses a mixture of experts to identify diverse contents for generating multiple target text~\citep{DBLP:conf/emnlp/ChoSH19}.

\noindent{\textbf{w/o z:}} our model without using mixture of experts.

\noindent{\textbf{w/o $\bm{\pi}$:}} our model without using soft templates.

\paragraph{Implementation Details} We set the number of experts $n_z$ to 3 and the number of soft templates $n_\pi$ to 12, for more values of $n_z$ and $n_\pi$. The embedding dimensions for the complexity level $d$, the expert $d_z$ and soft template $\pi$, are set to 30, 50 and 50 respectively.
We set hidden vector sizes to 256. Models are optimized with the Adam~\citep{kingma2015adam} and we initially set the learning rate to 0.001. Other standard parameters follow the default settings of the Pytorch\footnote{\url{https://pytorch.org}}.
We stop the training iterations until the performance difference between two consecutive iterations is smaller than 1e-6. 
For QG models that cannot be complexity-controllable, we concatenate the complexity vector with the hidden state from the encoder to initialize the decoder. 

\paragraph{Metrics} Automatic and human evaluation metrics are used to analyze the model's performance.
\textbf{Automatic Metrics:}
Following prior works~\citep{DBLP:conf/nlpcc/ZhouYWTBZ17,DBLP:conf/acl/PanXFCK20}, we use the metrics BLEU-4 (B-4) and ROUGE-L (R-L)~\citep{DBLP:journals/corr/abs-2006-14799} to evaluate the quality of generated questions against references. The generated questions might have different complexity levels than the input ones. Thus we also report F1-score (F1) based on the discrepancy between the complexity levels of generated questions labeled by our complexity estimator and the input ones.
\textbf{Human Metrics:}
We randomly select 200 pairs of passage and answer from the test datasets in HotpotQA and SQuAD respectively (400 cases in all), and manually evaluate the questions generated by all methods. 
Three annotators are asked to judge each question independently according to the following four criteria on the Likert scale of 1--5, with 1 being the worst and 5 being the best. \textbf{Naturalness (Nat.)} rates the fluency and comprehensibility of the generated question. 
\textbf{Complexity (Cpx.)} is used to measure the complexity of correctly answering a generated question in a given passage. The higher the complexity, the more difficult it is to find the answer.  
Given the same passage and answer, we expect questions generated by two different complexity levels to be distinct. Therefore, We employ \textbf{Diversity (Div.)} to measure the differences between the two questions with different complexity levels based on the same passage and answer.

\subsection{Results and Analysis for CCQG}
\label{results}
\textbf{Automatic Evaluation.}
Table~\ref{auto results on datasets} indicates the results of automatic evaluation, we can observe that:

1. For overall performance, our model achieves the best performance across all metrics. Specifically, our model improves the BLEU-4 and ROUGE-L by at least 1.16 and 1.31, respectively, over the best baseline DeepQG, which is specifically designed for generating complex questions. 

2. Our model achieves also superior consistency between input and output complexity levels in terms of F1 than all baselines, which use a single vector for each complexity label, attesting to our model's effectiveness in complexity modeling with mixture of experts and soft templates.

3. It is no surprise that generation of complex questions is more challenging. Our model and all baselines perform slightly better in terms of BLEU-4 and ROUGE-L on simple question generation than complex question generation. In contrast, complexity control is not always more difficult for some baselines on generating complex questions.

\noindent\textbf{Human Evaluation.} We conduct human evaluation to inspect if our findings of automatic evaluation are consistent with human perception. Apart from using the above mentioned metrics, we also provide sample questions generated by different models with varying complexity levels in Table \ref{case}.
\begin{table*}[htbp] 
   \centering
    \caption{Examples generated by our model and baselines, given the same passage and answer from HotpotQA.} 
    \label{case}
    \resizebox{\textwidth}{!}{
    \begin{tabular}{l|l} 
     \toprule
     \multirow{4}{*}{Passage} 
     &The 2013 Liqui Moly Bathurst 12 Hour was an endurance race for a variety of GT and touring car classes, including: GT3 cars, GT4 cars, Group 3E Series Production Cars and Dubai 24 Hour cars.\\
     &The event, which was staged at the Mount Panorama Circuit, near Bathurst, in New South Wales, Australia on 10 February 2013, was the eleventh running of the Bathurst 12 Hour.\\
     &Mount Panorama Circuit is a motor racing track located in Bathurst, New South Wales, Australia. The 6.213 km long track is technically a street circuit, and is a public road, with normal speed\\
     &restrictions, when no racing events are being run, and there are many residences which can only be accessed from the circuit.\\
      \midrule
    NQG++ & How long is the track? \textcolor{blue}{(simple)} 
     How long is the long track? \textcolor{blue}{(complex)}\\
    \midrule
     DLPH & What is the length of the track? \textcolor{blue}{(simple)} 
      What is the length of Mount Panorama Circuit? \textcolor{blue}{(complex)} \\
     \midrule
    DeepQG & What is the length of Mount Panorama Circuit, located in Bathurst, New South Wales? \textcolor{blue}{(simple)}  
      What is the length of the track, located in Bathurst, New South Wales, Australia? \textcolor{blue}{(complex)}\\
      \midrule
     MoE & How long is the track?  \textcolor{blue}{(simple)} 
      What is the length of the track? \textcolor{blue}{(complex)} \\
     \midrule
      CCQG & \textbf{How long is the track? \textcolor{blue}{(simple)} }
     \textbf{What is the length of the track at which the 2013 Liqui Moly Bathurst 12 Hour was staged? \textcolor{blue}{(complex)}} \\
    \midrule
       w/o z &What is the length of the track? \textcolor{blue}{(simple)} 
        What is the length of the track which is located in Bathurst, New South Wales? \textcolor{blue}{(complex)}  \\
          \midrule
         w/o $\pi$ & How long is the track?  \textcolor{blue}{(simple)} 
      What is the length of the track?  \textcolor{blue}{(complex)} \\
     \midrule
     \textbf{Gold}&\textbf{What is the length of the track where the 2013 Liqui Moly Bathurst 12 Hour was staged?} \textcolor{blue}{(complex)}\\
     \bottomrule
    \end{tabular} 
    }
 \end{table*}

1. \textbf{Naturalness} measures semantic and linguistic quality of generated questions. From Table \ref{human results on datasets} we can see that our model is superior in this metric in comparison to the SOTA models. Due to the task complexity, all models perform still better on simple questions than complex ones. As we can see from the sample questions in Table \ref{case}, the length of complex questions is relatively longer than that of the simple ones. Our close inspection also shows that our model generated more questions with complex syntactic structures than the baselines.

2. On simple question subsets, our model obtains the lowest \textbf{Complexity}, and conversely, on complex question subsets, we obtain the highest, which shows that our model is more capable of generating questions at the target complexity level. 

3. \textbf{Diversity} is measured between two questions of different complexity levels, given the same passage and answer. The results show that CCQG yields the highest diversity, which leads to the conclusion that MoE and soft templates make the generated questions with varying complexities more distinct from each other. In contrast, a single vector for each complexity level makes the baselines difficult to generate substantially diverse questions.

\noindent\textbf{Ablation Analysis.}
\label{Ablation Analysis}
To further investigate the effectiveness of the MoE and soft templates, we perform the experiments by removing them respectively.

\textbf{Effect of Expert $z$.}
From Tables~\ref{auto results on datasets} and~\ref{human results on datasets}, we can observe that the model (\textbf{w/o z}) performance drops obviously on complexity controlling (\textbf{F1} avd \textbf{Cpx.}) and diversity (\textbf{Div.}). We believe the main reason is that different experts $z$ captures different modes of the underlying distributions, thus effectively play the vital role for selecting template elements at the target complexity level.

\textbf{Effect of Soft Templates $\pi$.}
Without the soft templates, our model (w/o $\pi$) degenerates into the baseline MoE. It is evident from Tables~\ref{auto results on datasets} and~\ref{human results on datasets} that the result of the model \textbf{w/o $\pi$} is very close to that of MoE. Concretely, all the metric values drop significantly, especially those related to the quality of the question, such as \textbf{B-4} and \textbf{R-L}. This shows that soft templates $\pi$ play an important role in the full model. We believe that, on the one hand, $\pi$ guarantees the quality of the generated questions by providing additional constraints (cluster centroids for similar questions). On the other hand, since the constraint information is different with different inputs (different cluster centroids are selected), it guides the model to generate more diverse questions.

\subsection{Evaluation of Complexity Estimator}
We evaluate the efficiency of the proposed complexity estimator on HotpotQA (in-domain) and SQuAD (out-domain). The threshold is tuned on the training set of HotpotQA. We compare  our model with two baselines. The first one is \textbf{QA-sys}~\citep{gao2019difficulty}, which evaluates the complexity level of a question based on whether QA models can answer it or not. 
The second one is the BERT-based classifier utilizing unsupervised domain adaptation  (\textbf{UDA})~\citep{DBLP:conf/lrec/NishidaNSAT20}. 
\noindent\textbf{In-Domain Evaluation:} 
Only HotpotQA has ground truth complexity levels. In-domain evaluation is conducted on the questions labeled as $easy$ or $hard$ in the corresponding train, validation, and test datasets. 
\noindent\textbf{Out-Domain Evaluation:}
The out-domain evaluation is conducted on SQuAD, whose questions are not labelled with complexity levels. We randomly sample 200 questions and employ three annotators to give feedback individually on the complexity level of each question on a scale of 1--3, with 1 being \textit{simple}, 2 being \textit{uncertain} and 3 being \textit{complex}. Only when the results of two or more annotators are consistent, the label is regarded as the final complexity level of a question. We exclude the questions annotated with \textit{uncertain} and use the remaining 187 questions for testing. Furthermore, to verify the reliability of annotators, we conduct a \textit{Fleiss' kappa} test for each annotator’s result. To this end, the kappa coefficients are 0.796, 0.794 and 0.776, respectively.

\begin{table}[htb]
\centering
\caption{In-domain and Out-domain evaluations of complexity estimator on SQuAD and HotpotQA, \textbf{p.s.}, \textbf{p.c.}, \textbf{t.s.} and \textbf{t.c.} refer to predicted as simple/complex, and true simple/complex, respectively.}
  \label{estimator evaluation}
  \renewcommand\tabcolsep{1pt} 
\resizebox{\columnwidth}{!}{
  \begin{tabular}{lcc|cc|cc|cc|cc|cc}
    \toprule
    Dataset& \multicolumn{6}{c}{HotpotQA (In-domain)} & \multicolumn{6}{c}{SQuAD (Out-domain)} \\
     \cmidrule(lr){2-7} \cmidrule(lr){8-13} 
    Method
    &  \multicolumn{2}{c}{QA-sys}     &  \multicolumn{2}{c}{UDA}  & 
    \multicolumn{2}{c}{Ours} & \multicolumn{2}{c}{QA-sys}    &   \multicolumn{2}{c}{UDA} &   \multicolumn{2}{c}{Ours}  \\
    \midrule
    & p.s. & p.c. & p.s. & p.c. & p.s. & p.c. & p.s. & p.c. & p.s. & p.c. & p.s. & p.c. \\
    \midrule
   t.s.  & 4,369 & 1,057 & 4,628 & 798 & 5,271 & 155 & 87 & 22& 76& 34& 93& 15  \\
   \midrule
   t.c &1,219 &2,398& 961& 2,656& 210& 3,407& 24& 54& 30& 47& 12& 67\\ 
   \midrule
   F1    &  \multicolumn{2}{c}{0.736}& \multicolumn{2}{c}{0.795}& \multicolumn{2}{c}{\textbf{0.958}}& \multicolumn{2}{c}{0.753}& \multicolumn{2}{c}{0.658}& \multicolumn{2}{c}{\textbf{0.856}} \\
    \bottomrule
  \end{tabular}
  }
\end{table}

\noindent\textbf{Results:} Table~\ref{estimator evaluation} reports F1-score and the confusion matrix for each method on the two datasets. (1) In the in-domain setting, our cross-domain estimator outperforms QA-sys and UDA in terms of F1 scores with a wide margin. QA-sys falls short of UDA by 5\%, which shows that it is not reliable to use the correctness of answering questions as a way of assessing complexity levels. (2) In the out-domain setting, QA-sys surprisingly achieves comparable performance in both settings, but is still more than 10\% behind our model. We conjecture that the relatively poor performance of both learning-based deep models may attribute to the domain specific spurious features that are irrelevant to complexity levels of questions.

\section{Discussion on MoE-based Architecture}
We provide justifications of the MoE-based architecture from the perspective of high-level cognition. 
Humans can easily ask questions that are simple and complex questions~\citep{DBLP:conf/nips/RotheLG17}, mainly because we can identify patterns through a certain mechanism and then combine these patterns for generalizing to various scenarios. 
That is, humans possess the capability for compositional generalization, which is critical for learning in real-world situations~\citep{DBLP:conf/nips/AtzmonKSC20}.
Some studies have shown the importance of modularity for this capability~\citep{sternberg2011modular,DBLP:journals/corr/abs-1207-2743}.
They suggest that modularity is conducive to the specialization of different modules, which are responsible for different functions. In other words, specialization improves generalization. 
Similarly, a modular neural network enables compositional generalization like human intelligence.
MoE-based architecture can be regarded as an implementation of this concept. MoE is a tightly coupled modular structure designed so that similar inputs are mapped to similar expert modules, effectively making each module specialize in a different selection.

\section{Conclusion and Future Work}
We propose a novel encoder-decoder model incorporating soft templates and MoE to address the problem of complexity-controllable question generation. As most domains do not have training data for \ccqg models, we propose a simple and effective cross-domain estimator to predict the missing complexity levels of questions. In the extensive experiments of both \ccqg and complexity assessment tasks, our models achieve superior performance over the competitive baselines across all experimental settings.
In the future, we will consider anaphora resolution and numerical reasoning in complexity estimator, and explore the performance of our model in different applications, such as examination and assisting QA systems.

\newpage
\section*{Acknowledgments}
This work was supported by the National Key Research and Development Program of China under grants [2017YFB1002801, 2018YFC0830200]; the Natural Science Foundation of China grants [U1736204]; the Fundamental Research Funds for the Central Universities [2242021k10011].
\bibliography{anthology,custom}
\bibliographystyle{acl_natbib}

\end{document}